\documentclass{article}

\usepackage{arxiv}

\usepackage[utf8]{inputenc} 
\usepackage[T1]{fontenc}    
\usepackage{hyperref}       
\usepackage{url}            
\usepackage{booktabs}       
\usepackage{amsfonts}       
\usepackage{nicefrac}       
\usepackage{microtype}      
\usepackage{lipsum}		
\usepackage{graphicx}
\usepackage{natbib}
\usepackage{doi}
\usepackage{calc}
\usepackage{indentfirst}
\usepackage{fancyhdr}
\usepackage{graphicx,epstopdf}
\usepackage{lastpage}
\usepackage{ifthen}
\usepackage{float}
\usepackage{amsmath}
\usepackage{amssymb} 
\usepackage[right]{lineno}
\usepackage{setspace}
\usepackage{enumitem}
\usepackage{mathpazo}
\usepackage{booktabs} 
\usepackage{titlesec}
\usepackage{etoolbox} 
\usepackage{tabto} 
\usepackage{xcolor, colortbl} 
\usepackage{soul} 
\usepackage{multirow}
\usepackage{microtype} 
\usepackage{tikz} 
\usepackage{refcount} 
\usepackage{changepage} 
\usepackage{attrib} 
\usepackage{upgreek} 
\usepackage{array} 
\usepackage{tabularx}
\usepackage{pbox} 
\usepackage{ragged2e} 
\usepackage[]{tocloft} 
\usepackage{marginnote} 
\usepackage{marginfix} 
\usepackage{enotez} 
\usepackage{xstring} 
\usepackage{listings}
\usepackage{cuted}
\usepackage{subcaption}
\usepackage{soul}
\usepackage{threeparttable}
\usepackage{tabularx}
\usepackage{caption}

\title{Connecting Vision and Emissions: A Behavioural AI Approach to Carbon Estimation in Road Design}

\author{ \href{https://orcid.org/0000-0003-1806-1189}{\includegraphics[scale=0.06]{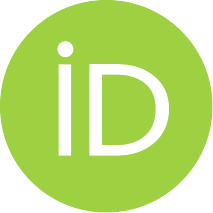}\hspace{1mm}Ammar K~Al Mhdawi}
	\And
	\href{https://orcid.org/0000-0002-5064-2621}{\includegraphics[scale=0.06]{orcid.pdf}\hspace{1mm}Nonso Nnamoko} 
	\And
	\href{https://orcid.org/0000-0003-1600-8587}{\includegraphics[scale=0.06]{orcid.pdf}\hspace{1mm}Safanah Mudheher Raafat} 
	\And
	\href{https://orcid.org/0000-0001-5870-0323}{\includegraphics[scale=0.06]{orcid.pdf}\hspace{1mm}M.K.S. Al-Mhdawi} 
	\And
	\href{https://orcid.org/0000-0000-0000-0000}{\includegraphics[scale=0.06]{orcid.pdf}
    \hspace{1mm}Amjad J~Humaidi} 
}

\renewcommand{\shorttitle}{Connecting Vision and Emissions}

\hypersetup{
pdftitle={Connecting Vision and Emissions: A Behavioural AI Approach to Carbon Estimation in Road Design},
pdfsubject={cs.AI, cs.LG},
pdfauthor={Ammar K~Al Mhdawi, Nonso Nnamoko},
pdfkeywords={Carbon emission, Behavioural AI, Image processing, OCR, YOLO, Vehicle detection},
}

\begin{document}
\maketitle

\begin{abstract}
	In this paper, we present an enhanced YOLOv8 real-time vehicle detection and classification framework, named YOLOv8-Deep, designed specifically for estimating carbon emissions in urban highway environments. The system enhances  the state-of-the-art YOLOv8 architecture to accurately detect, segment, and track vehicles from live traffic video streams. Once a vehicle is localized, a dedicated deep learning-based identification module is employed to recognize license plates and classify vehicle types—an essential step, given that carbon emission rates vary significantly across different vehicle categories.
    While YOLOv8 is highly effective in general-purpose object detection, it lacks the built-in capacity for fine-grained recognition tasks such as reading license plates or determining vehicle attributes beyond class labels. To address this, our framework incorporates a hybrid pipeline where each detected vehicle is tracked and its bounding box is cropped and passed to a deep Optical Character Recognition (OCR) module. This OCR system, composed of multiple convolutional neural network (CNN) layers, is trained specifically for character-level detection and license plate decoding under varied conditions such as motion blur, occlusion, and diverse font styles.
    Additionally, the recognized plate information is validated using a real-time API that cross-references with an external vehicle registration database to ensure accurate classification and emission estimation. This multi-stage approach enables precise, automated calculation of per-vehicle carbon emissions using standardized emission factors. Extensive experimental evaluation was conducted using a diverse vehicle dataset enriched with segmentation masks and annotated license plates. The YOLOv8 detector achieved a mean Average Precision (mAP\@0.5) of approximately 0.71\% for bounding boxes and 0.70\% for segmentation masks. Character-level OCR accuracy reached up to 99\% with the best-performing CNN model. The system’s training performance was visualized through precision-recall curves, F1-score trends, loss evolution plots, and confusion matrices, all of which demonstrate consistent learning progression and highlight key areas for refinement.
    These results affirm the feasibility of combining real-time object detection with deep OCR for practical deployment in smart transportation systems, offering a scalable solution for automated, vehicle-specific carbon emission monitoring.
\end{abstract}


\section{Introduction}
\label{sec:introduction}

The growing urgency to address climate change has intensified research efforts aimed at reducing greenhouse gas emissions, with the transportation sector recognized as a major contributor to global CO$_2$ output \cite{Favelli_Xie_Tonoli_2024}. Traditional methods for monitoring vehicle emissions—such as periodic inspections, laboratory testing, or sensor-based systems—often fall short in capturing the dynamic, real-world behavior of individual vehicles \cite{Dikshit_Atiq_Shahid_Dwivedi_Thusu_2023}. Consequently, there is increasing interest in developing intelligent, real-time systems that can provide granular insights into vehicular emissions to inform sustainable urban planning and policy-making. Advances in artificial intelligence (AI), computer vision, and edge computing have already shown promise in various domains of activity monitoring and environmental surveillance \cite{Trabelsi_Khemmar_Decoux_Ertaud_Butteau_2022, Jha_Marzban_Hu_Mahmoud_Al‐Dhahir_Busso_2021}.

Numerous studies have explored ways to estimate vehicle emissions. Direct measurement approaches, such as those using roadside infrared/UV spectroscopy \cite{Patino-Aroca2022-oe}, static air quality sensors \cite{Rodriguez_Valido2022-lp}, or onboard diagnostics (OBD) \cite{rivera-Campoverde2021-se}, provide valuable data but face significant limitations. Sensor-based methods are often costly and lack the ability to attribute emissions to individual vehicles in mixed traffic, while OBD solutions are constrained to instrumented vehicles only. To overcome these hardware dependencies, alternative strategies like traffic simulation and fleet-wide emission models have been proposed \cite{Cai2018-qu, Saija2002-od}. Although scalable, such approaches depend heavily on assumptions and cannot adapt to real-time traffic variations.

Parallel to this, computer vision techniques—particularly deep learning models like YOLO—have demonstrated high accuracy in real-time vehicle detection and classification tasks \cite{Zhang2024-vx, Yavuz2024-ba}. These advances open possibilities for image-based, non-intrusive monitoring of urban environments. Yet, despite their success in vehicle detection, such systems have not been leveraged to directly estimate carbon emissions at the individual vehicle level. The closest related study integrated roadside cameras with air quality sensors to correlate traffic activity with ambient CO$_2$ levels \cite{Rodriguez_Valido2022-lp}, but did not attribute emissions to specific vehicles nor provide per-vehicle estimates.

Despite the extensive body of work on emissions monitoring and vehicle detection, no existing solution integrates real-time image-based vehicle recognition with dynamic CO$_2$ estimation at the level of individual vehicles. This gap highlights an opportunity to develop a scalable, non-invasive system that not only detects vehicles using computer vision but also infers their potential emissions using classification outputs. Such a system could mitigate reliance on hardware-intensive or fleet-specific techniques and improve emission tracking granularity in dense urban settings.

This research introduces a novel vision-based framework for estimating vehicle-specific CO$_2$ emissions in real-time using behavioral AI. The core of our method is an enhanced YOLOv8 model, modified with a final convolutional layer for precise vehicle license plate detection and segmentation. The model uses a road traffic dataset from Roboflow~\cite{mmsseg-dataset} to detect vehicle type and a number plate dataset~\cite{number-plates-dataset} (also from Roboflow) for plate recognition. The latter allows for vehicle search on a dummy Application Programming Interface (API) simulated using Beeceptor\footnote{www.beeceptor.com} to obtain CO$_2$ characteristics of the detected vehicle. The API is designed to mimic the Driver and Vehicle Licensing Agency (DVLA) Vehicle Enquiry API~\cite{dvla2025-ab} which is not publicly available. By classifying vehicle types and linking them to emission factor databases, our system infers likely emissions per vehicle directly from traffic camera feeds. This approach bridges the gap between visual traffic monitoring and carbon tracking by offering a scalable, automated pipeline without the need for expensive physical infrastructure.

The key contributions of this work are as follows:

\begin{itemize}
\item We develop an enhanced YOLOv8-based deep learning model capable of real-time vehicle and license plate detection to support fine-grained classification.
\item We propose an automated carbon emission estimation framework that integrates vision-based classification with emission factor mapping in a real-time API environment.
\item We demonstrate the feasibility of a scalable, infrastructure-light solution for individual vehicle CO$_2$ emission inference from live traffic footage.
\end{itemize}

The rest of the paper is structured as follows: Section~\ref{sec:relatedWork}  provides details about related work and the necessary background legislation underpinning the study direction. The experimental data and methodology approach, including details about the experiment setup and evaluation measures, are presented in Section~\ref{sec:methods}. Section~\ref{sec:results} presents our findings including result analysis and any issues likely to threaten the validity of results. Section~\ref{sec:conclusion} summarises the study and points out future work.

\section{Related Work} \label{sec:relatedWork}

This section provides an overview of the regulatory context and technical foundations relevant to estimating vehicle emissions using image processing techniques. It begins by outlining key environmental legislation that motivates the need for scalable and accurate emissions monitoring. It then reviews existing work on vehicle detection and classification from video streams, followed by studies that link vehicle identification to emission estimation. Finally, the section highlights a gap in the literature where emissions are inferred directly from visual classification without relying on intrusive methods.

\subsection{Environmental Legislation}

Efforts to reduce carbon emissions from transport have been underpinned by increasingly stringent environmental regulations. In the United Kingdom (UK), the Climate Change Act 2008 legally commits the government to reduce greenhouse gas emissions to net zero by 2050~\cite{ukclimate2008}. At the European level, Regulation (EU) 2019/631 sets CO$_2$ emission performance standards for new passenger cars and light commercial vehicles, incentivising cleaner vehicle production and the gradual phase-out of high-emitting models~\cite{euregulation2019}. These regulations necessitate accurate and scalable methods to monitor real-world emissions, especially as concerns mount over discrepancies between laboratory test results and real-world driving emissions~\cite{franco2014real}.

Traditional approaches to monitoring vehicle emissions typically rely on roadside sensors~\cite{ren2022-nk} or onboard diagnostics (OBD)~\cite{rivera-Campoverde2021-se}. However, these methods can be costly, intrusive, or restricted in scope. For instance, while OBD provides accurate emission data for individual vehicles, it cannot be accessed externally. As a result, there is increasing interest in using other non-intrusive alternatives for assessing vehicle emissions.

\subsection{Image-based Vehicle Detection \& Classification}

Image processing and machine learning techniques have been widely adopted to detect and classify vehicles from video footage. CNNs in particular have achieved strong performance in vehicle detection tasks~\cite{bertozzi1997-oj,azimjonov2021-gc,bie2023-hm,anandhalli2018-qg}. Datasets such as UA-DETRAC~\cite{wen2020ua} and CityFlow~\cite{tang2019cityflow} have facilitated the development of robust models capable of classifying vehicle types (e.g., sedan, SUV, truck) in varied lighting and weather conditions. while once valuable benchmarks for vehicle detection and tracking, both datasets are now less commonly used in recent object detection research, particularly with modern You Only Look Once (YOLO)~\cite{redmon2016yolo} variants. This shift is largely due to their limited diversity, dated scenarios, and narrow focus on traffic environments. In contrast, newer datasets such as COCO-Traffic~\cite{david2021-ab}, BDD100K~\cite{yu2020-fc}, nuScenes~\cite{caesar2020-nu}, Waymo~\cite{sun2020-wm} and MOTChallenge~\cite{milan2016-mo} provide richer annotations, greater environmental diversity, and support for multi-task learning. However, none of these were directly suitable for the task presented in this study. Thus, we utilised a road traffic dataset available within a Roboflow project~\cite{mmsseg-dataset} which aligns more closely with our research priorities to detect vehicle type, making it the preferred choice for the additional emission research presented. Methods applied on the dataset typically involve object detection frameworks such as YOLO or Faster R-CNN~\cite{ren2015faster}, which are trained to localize and classify vehicles within frames. Several studies have also employed transfer learning from large-scale datasets to improve performance on domain-specific tasks like traffic surveillance~\cite{liu2020vehicle}.

\subsection{Estimating CO\textsubscript{2} from Vehicles}

Several studies have estimated vehicle CO$_2$ emissions using direct measurement techniques, such as roadside sensors (e.g., remote sensing, IoT air quality monitors) and onboard diagnostics (OBD). For instance, Pati{\~n}o-Aroca et al.~\cite{Patino-Aroca2022-oe} employed infrared/UV spectroscopy to measure emissions from passing vehicles, while Valido et al.~\cite{Rodriguez_Valido2022-lp} used static MQ-135 sensors to monitor ambient CO$_2$ near roadways. Similarly, OBD-based methods~\cite{rivera-Campoverde2021-se} extract real-time engine data to compute emissions. However, these approaches face critical limitations: (1) sensor-based systems (e.g., remote sensing, IoT) are expensive to deploy at scale and cannot attribute emissions to specific vehicles in mixed traffic, and (2) OBD data is restricted to instrumented vehicles, making it inaccessible for fleet-wide analysis.

To overcome hardware dependencies, some studies have adopted traffic simulation and macro-level emission models. For example, Cai et al.~\cite{Cai2018-qu} used traffic volume data and bottom-up modeling to estimate urban CO$_2$ emissions, while Saija et al.~\cite{Saija2002-od} derived static emission factors based on fleet composition. Though scalable, these methods lack real-time granularity and rely on theoretical assumptions that may not reflect dynamic traffic conditions.

Recent advances in computer vision have enabled vehicle detection and classification using surveillance cameras (e.g., YOLOv7 in Zhang et al.~\cite{Zhang2024-vx} and edge-computing in Yavuz et al.~\cite{Yavuz2024-ba}. The most closely related work~\cite{Rodriguez_Valido2022-lp} combined roadside cameras with low-cost sensors to correlate traffic activity with localized CO$_2$ levels. However, this study did not estimate emissions per vehicle and instead measured aggregate ambient air quality, failing to link detected vehicles to their specific emissions. This creates an opportunity to extend research in this area to account for individual vehicle emissions.

\subsection{Research Gap}

While prior works have explored sensor-based measurements, traffic modeling, or standalone vehicle detection, none integrate real-time vision-based vehicle detection with dynamic CO$_2$ computation at the individual vehicle level. To the best of our knowledge, no existing work has demonstrated a pipeline that estimates potential carbon emissions using only image-based classification of vehicles from traffic video streams. This presents an opportunity to explore a scalable method for emissions monitoring that can supplement or enhance current systems. Our approach addresses this by (1) eliminating dependency on costly roadside hardware, (2) enabling fleet-wide analysis without OBD restrictions, and (3) providing granular, real-time emission estimates that macro-models cannot achieve.

\section{Materials and Methods} \label{sec:methods}

This section outlines the methodological approach adopted for estimating vehicle emissions from video streams. Section~\ref{sec:design} presents the design of the proposed pipeline, which integrates vehicle detection and license plate recognition with external API queries to retrieve vehicle-specific attributes—such as CO$_2$ classification—for emission estimation along a defined road segment. Section~\ref{sec:experimentSetup} describes the experimental setup, including the technical tools, libraries, and datasets employed to implement and evaluate the pipeline.

\subsection{framework design} \label{sec:design}

\begin{figure*}[h]
    \centering
    \includegraphics[width=0.8\linewidth]{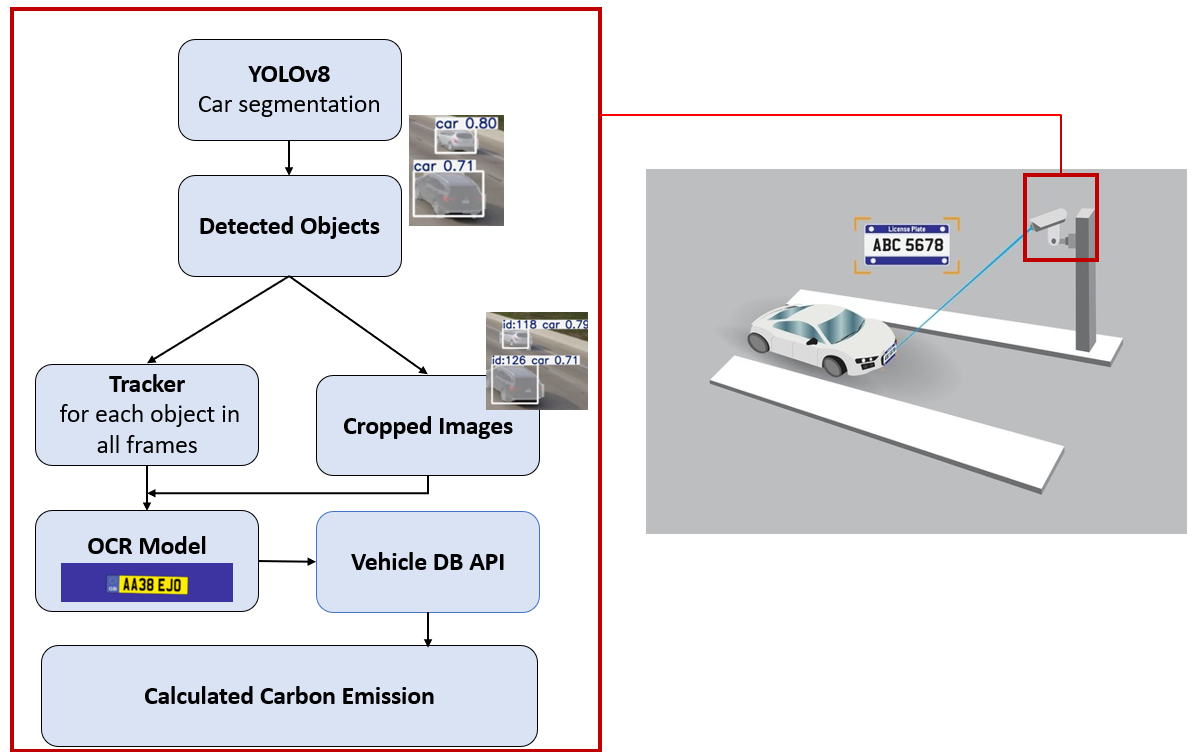} 
    \caption{Framework Pipeline: An end-to-end system for vehicle detection, license plate recognition, vehicle information retrieval, and carbon emission estimation.}
    \label{fig:framework_pipeline}
\end{figure*}

The proposed framework is structured as a multi-stage pipeline designed for real-time vehicle detection, license plate recognition, and carbon emission estimation in urban traffic environments as shown in Figure~\ref{fig:framework_pipeline}. Each component of the system performs a specific function that collectively contributes to identifying vehicles and calculating their environmental impact. The process begins with the YOLOv8 model, a state-of-the-art object detection architecture that detects and segments vehicles within video frames. YOLOv8 outputs both bounding boxes and segmentation masks, enabling accurate localization and outlining of vehicle shapes. This segmentation enhances the precision of downstream processing. Following detection and segmentation, a tracking and cropping module is applied to the segmented vehicle objects. This module tracks each vehicle across successive video frames using a unique identifier, ensuring that every vehicle is counted and processed only once. The cropped images of these tracked vehicles are then passed to the next stage of the system. These cropped vehicle images are fed into a deep Optical Character Recognition (OCR) model, which is trained to identify and extract license plate numbers. The OCR system handles a variety of real-world challenges such as motion blur, poor lighting, and distorted characters by using convolutional and recurrent neural network layers. The output of this model is the decoded license plate number for each tracked vehicle. The recognized license plate numbers are then used to query a remote vehicle registration API. This external service returns detailed information about each vehicle, including its classification (such as private car, truck, or taxi), fuel type, engine size, and an associated carbon emission value based on standardized emission factors. Using the vehicle type and emission rate provided by the API, the system calculates the estimated carbon emission for each vehicle. This calculation also takes into account the duration or distance a vehicle appears in the video frame. By aggregating this information across multiple vehicles and timeframes, the system provides a reliable estimate of total traffic-related carbon emissions within the monitored area. This integrated approach allows for a real-time, automated, and accurate analysis of vehicle behaviour and associated environmental impact, making it a valuable tool for urban planning, traffic monitoring, and sustainability initiatives.
The research was evaluated using standard performance metrics, including precision, recall, mean Average Precision (mAP), and multiple loss components, namely: bounding box, classification, segmentation, and distributional focal loss. These metrics were monitored throughout training to assess the model's convergence behavior and its generalization capability on unseen validation data.

\subsection{Experimental Setup} \label{sec:experimentSetup}

In our experimental setup,  the process begins with dataset preparation, using two primary sources. The first is a locally uploaded ZIP file of the plate number dataset~\cite{number-plates-dataset} (named \texttt{plates.zip}), which is extracted into the working directory and contains a small sample of vehicle license plate images used for testing the OCR component. The second dataset downloaded from Roboflow~\cite{mmsseg-dataset}, was formatted for YOLOv8-OBB (oriented bounding boxes), designed for more precise object detection in traffic scenes. Although the exact number of images is not specified, it includes annotations suitable for vehicle segmentation and detection.

The deep learning model used in this setup is \texttt{yolov8n-seg.pt}, a lightweight segmentation model from the YOLOv8 family by Ultralytics\footnote{www.ultralytics.com}. This model is pretrained on a broader dataset but is fine-tuned in this research for vehicle segmentation tasks using the downloaded traffic dataset. Training is conducted for 20 epochs with a batch size of 12 and an image size of $640\times640$ pixels. Only one worker thread was used for loading the data during training. This configuration strikes a balance between training efficiency and system resource usage, suitable for real-time applications.

Once the model is trained, it is used to process video footage—specifically a file named \texttt{highway-test.mp4}. The model segments and detects vehicles in each frame of the video. YOLOv8’s tracking module is then used to assign persistent ids to each detected object across frames using the in-built \texttt{ByteTrack} tracker configuration. This ensures that each unique vehicle is counted only once, regardless of how many frames it appears in. The code includes logic to map raw class labels (e.g., "suv", "van", "pickup") into broader categories such as "car", "truck", "motorcycle", and "bus". These mappings simplify the final analysis and help consolidate similar vehicle types under unified labels. Unique object ids are tracked per category using a dictionary, allowing for accurate vehicle counting without duplication.

For license plate recognition, the \texttt{EasyOCR} library was employed. It processes both still images and video frames to extract text data from vehicle license plates. OCR is performed on a sample image and on each frame of the tracked video output. Detected texts are filtered to exclude short or irrelevant entries, with valid plates expected to be at least 6 characters long and ideally 7 characters to align with standard formats. These extracted plate numbers are then passed to a mock DVLA API, simulated using Beeceptor. This API accepts a plate number and returns a dummy JSON response containing vehicle information such as make, model, registration details and CO$_2$ class. The system is designed to display this information in a readable format, demonstrating how such a pipeline could integrate with a real vehicle database. Table~\ref{tab:setup} shows a snapshot of the technical tools and libraries used in the pipeline.

\begin{table}[h]
\centering
\caption{Experimental setup components} \label{tab:setup}
\begin{tabular}{ll}
\toprule
\textbf{Component} & \textbf{Details} \\
\midrule
Dataset 1 & \texttt{plates.zip} extracted to \texttt{G drive /content} \\
Dataset 2 & Road-Traffic-4 (Roboflow v4), YOLOv8-OBB format \\
YOLO Model & Pre-trained \texttt{yolov8n-seg.pt} (segmentation) \\
Training Params & 20 epochs; batch 12; image 640; workers 1 \\
Tracking & ByteTrack via \texttt{model.track()} \\
OCR Tool & EasyOCR for plate text extraction \\
Video Source & \texttt{highway-test.mp4} \\
Vehicle API & Beeceptor mock DVLA endpoint \\
\bottomrule
\end{tabular}
\end{table}

Additional configurations include setting the detection confidence threshold to 0.25 and the intersection over union (IOU) threshold to 0.45 to balance precision and recall during detection. The experiment also includes a utility function for zipping output folders, likely used to archive training results. Overall, this experimental setup forms a complete end-to-end pipeline that combines object detection, segmentation, tracking, and OCR to analyze road traffic and vehicle data from video, with the potential to scale or adapt for real-time traffic monitoring applications. Table 2 is based on data reported by regulatory agencies such as the U.S. Environmental Protection Agency~\cite{epa2023} and the European Environment Agency~\cite{eea2024}, which provide annual statistics on emissions performance across various car segments. These figures offer insight into how vehicle design and energy source affect environmental impact, and they underscore the critical role of electrification and hybridization in reducing transport-sector emissions.


\begin{table}[h!]
\centering
\caption{Avg. CO\textsubscript{2} Emissions by Car  \cite{epa2023} \cite{eea2024}}
\label{tab:emissions}
\begin{threeparttable}
\begin{tabular}{lcc}
\toprule
\textbf{Class} & \textbf{Fuel} & \textbf{CO\textsubscript{2} (g/km)} \\
\midrule
Subcompact     & Gasoline      & 115 \\
Compact        & Gas/Diesel    & 125 \\
Midsize        & Gas/Diesel    & 140 \\
Full-size      & Gas/Diesel    & 160 \\
SUV            & Gas/Diesel    & 180 \\
Pickup         & Gas/Diesel    & 200 \\
Luxury         & Gasoline      & 170 \\
Electric       & Electric      & 0   \\
Hybrid         & Hybrid        & 90  \\
\bottomrule
\end{tabular}
\begin{tablenotes}
\small
\item \textit{Note: Higher emissions in large vehicles; EVs produce no tailpipe CO\textsubscript{2}.}
\end{tablenotes}
\end{threeparttable}
\end{table}


\section{Results and Analysis} \label{sec:results}
This section presents the experimental results obtained after extensive training and evaluation of the proposed YOLOv8-based detection and segmentation model. The training process involved both object detection and instance segmentation tasks, applied to a custom dataset containing various vehicle classes such as buses, taxis, trucks, government vehicles, and private cars. In addition to vehicle detection, license plate recognition was also integrated to enable detailed classification and support downstream tasks such as carbon emission estimation. The dataset used was carefully curated to reflect real-world urban traffic conditions, incorporating diverse lighting, angles, occlusions, and scene complexities. Each image was annotated with bounding boxes and segmentation masks, and many included labeled license plates to support fine-grained recognition. Following model training over multiple epochs, the YOLOv8 system demonstrated consistent improvement in key metrics such as precision, recall, mean Average Precision (mAP), and multiple loss components (box, classification, segmentation, and distributional focal loss). These metrics were tracked to understand the model’s convergence behavior and its ability to generalize to unseen validation data. Beyond detection, the workflow included a post-processing phase in which the predicted bounding boxes were used to isolate individual vehicles in the scene. Each bounding box was tracked and cropped from the original image frames, producing isolated vehicle patches. These cropped images were then fed into a deep OCR (Optical Character Recognition) model, specifically designed to extract alphanumeric text from license plates. Moreover, the OCR system consisted of a multi-layer convolutional neural network trained on diverse license plate formats and font styles. This enabled accurate character recognition even in challenging conditions such as blur, partial occlusion, or non-standard fonts. By combining YOLOv8’s localization capabilities with deep OCR-based recognition, we were able to detect, extract, and interpret license plate numbers in real time, forming a crucial part of the system’s end-to-end vehicle identification pipeline. In summary, these results are further visualized through plots showing training and validation performance trends, confusion matrices, and distribution analyses, offering a comprehensive understanding of the system’s strengths and areas for optimization.

\begin{figure*}[h]
\begin{center}
   
    \centering
    \begin{subfigure}[b]{0.48\textwidth}
        \centering
        \includegraphics[width=\textwidth]{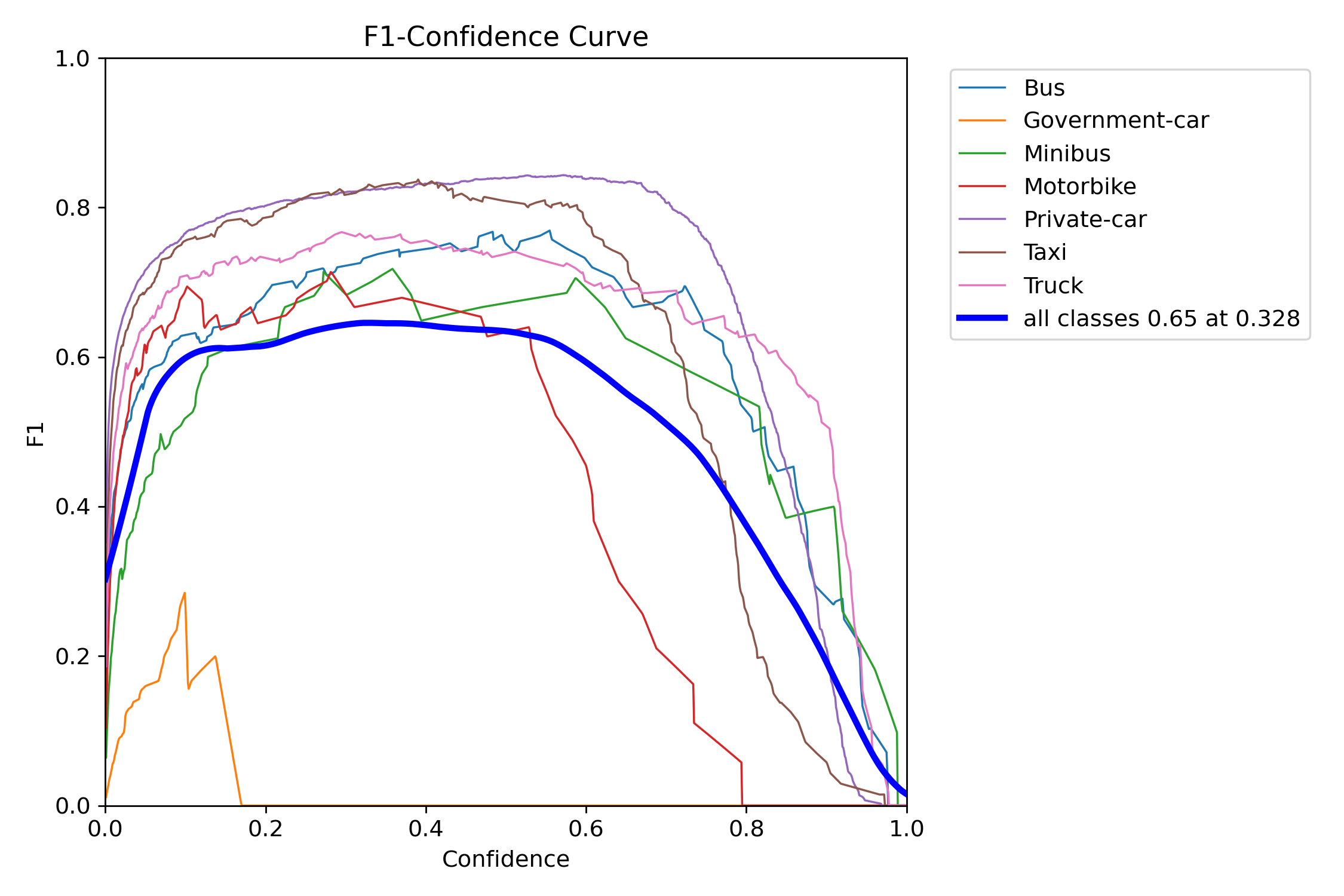}
        \caption{F1-Confidence Curve (Mask)}\label{fig:MaskF1_curve}
    \end{subfigure} 
    \hfill 
    \begin{subfigure}[b]{0.48\textwidth}
        \centering
        \includegraphics[width=\textwidth]{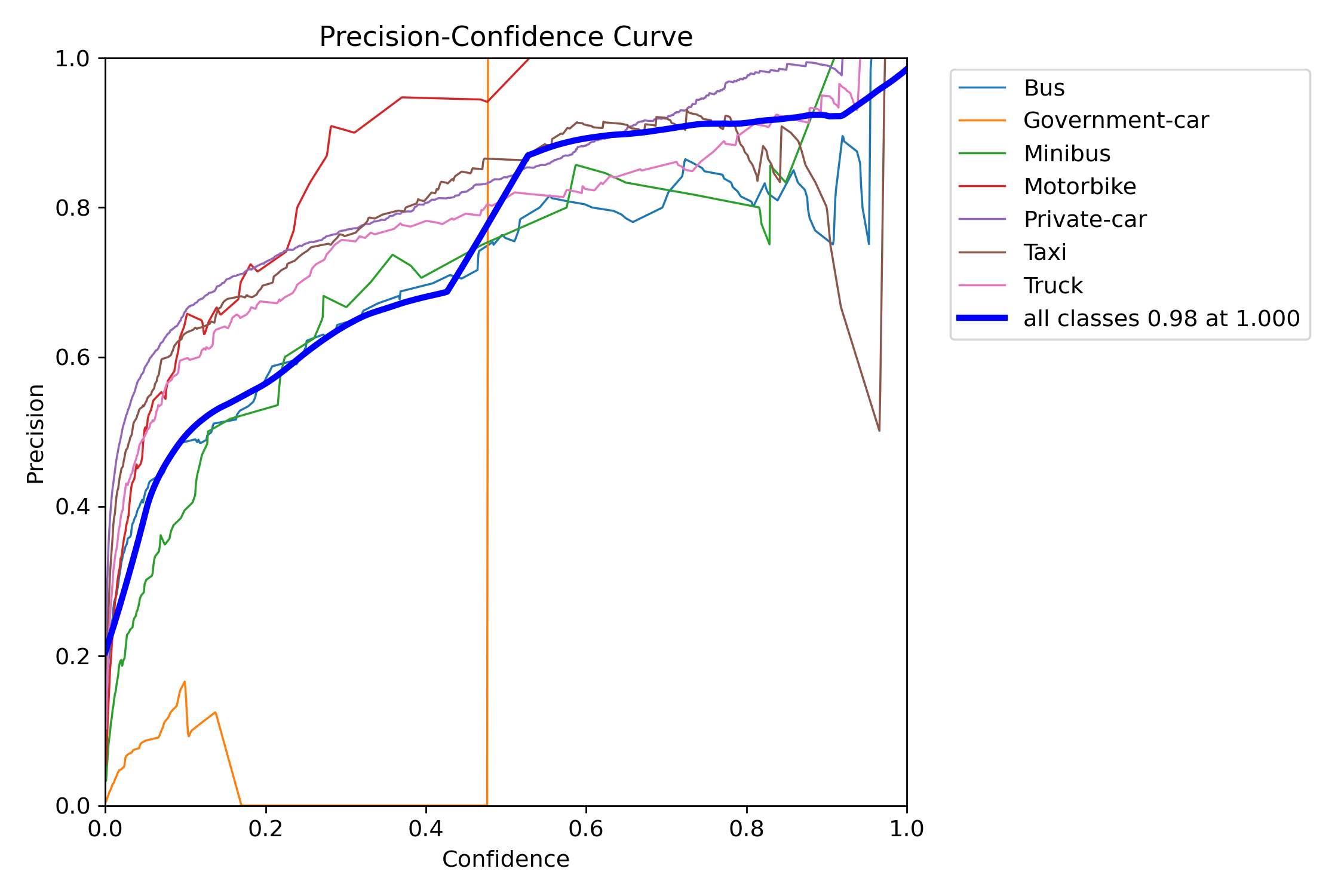}
        \caption{Precision-Confidence Curve (Mask)}\label{fig:MaskP_curve}
    \end{subfigure}
    \vspace{0.5cm}
    \begin{subfigure}[b]{0.48\textwidth}
        \centering
        \includegraphics[width=\textwidth]{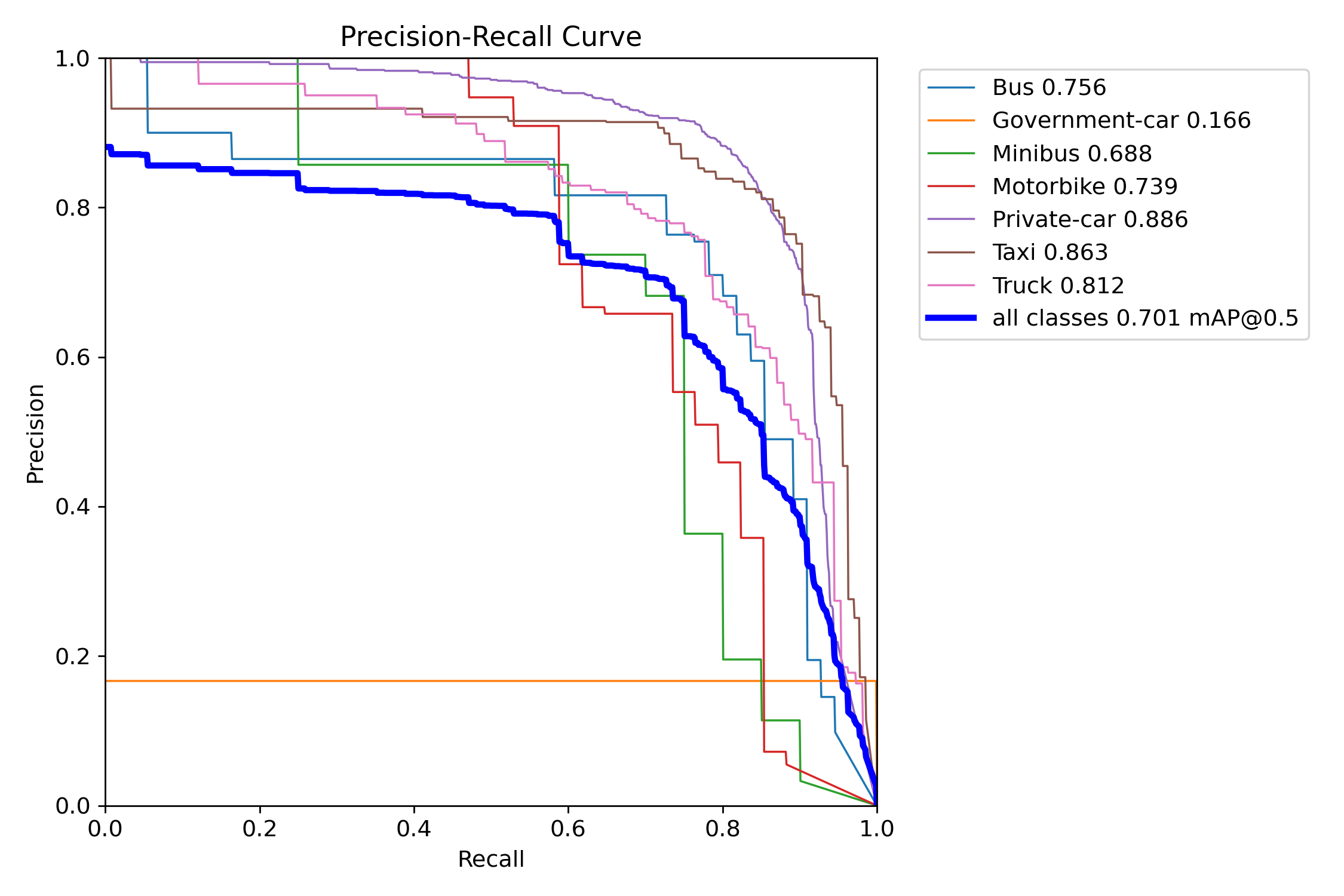}
        \caption{Precision-Recall Curve (Mask)}\label{fig:MaskPR_curve}
    \end{subfigure}
    \hfill 
    \begin{subfigure}[b]{0.48\textwidth}
        \centering
        \includegraphics[width=\textwidth]{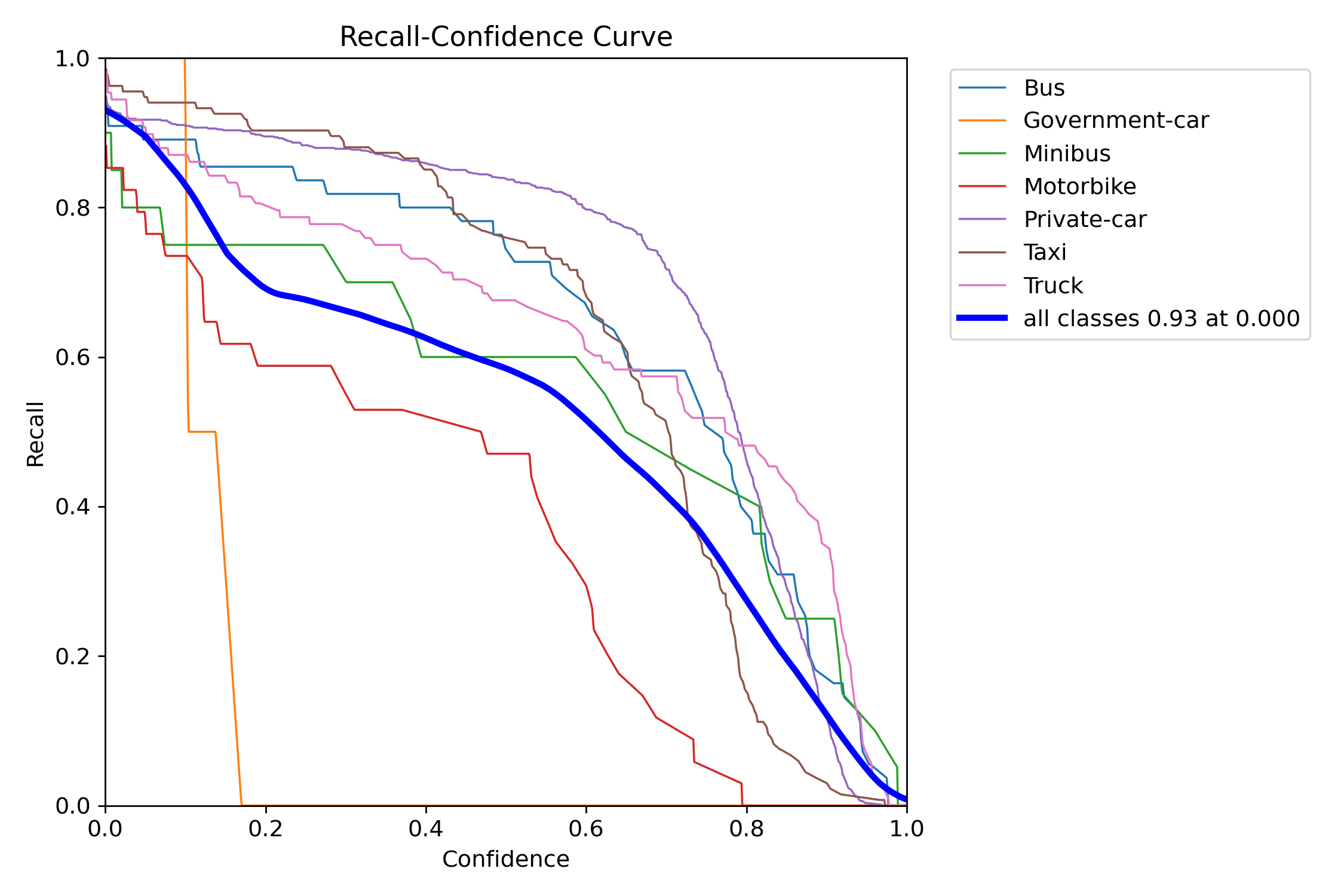}
        \caption{Recall-Confidence Curve (Mask)}\label{fig:MaskR_curve}
    \end{subfigure}
    \caption{Comparison of various curves for the model (Mask). Figure~\ref{fig:MaskF1_curve} shows the F1 score variation across confidence thresholds for each vehicle class using segmentation masks. The optimal threshold for all classes is at 0.328, where F1 reaches 0.65; Figure~\ref{fig:MaskP_curve} shows precision versus confidence. The model achieves its highest precision of $0.98$ at the confidence threshold of $1.000$ for all classes; Figure~\ref{fig:MaskPR_curve} shows the trade-off between precision and recall using mask outputs. The average precision (mAP@0.5) across all classes is 0.701; and Figure~\ref{fig:MaskR_curve} shows how recall changes with increasing confidence. The model achieves maximum recall of 0.93 at a threshold of 0.}

\end{center}
\end{figure*}





Figure \ref{fig:MaskF1_curve} presents how the F1-score (the harmonic mean of precision and recall) changes with varying confidence thresholds for each class. It reveals that most classes achieve high F1-scores between confidence levels of 0.2 and 0.5, with the overall maximum F1-score (0.65) occurring at a confidence threshold of 0.328. The segmentation masks provide fine-grained region localization, leading to more robust F1 values across classes such as Private-car, Taxi, and Truck. The Government-car class performs poorly, likely due to class imbalance or limited examples.

Figure \ref{fig:MaskP_curve} shows the relationship between confidence threshold and precision. Precision improves as the confidence increases, peaking at 1.0 with 0.98 precision. This indicates that high-confidence predictions are more likely to be accurate, although fewer predictions are made. The red and brown lines (Motorbike and Taxi) demonstrate strong performance, while Government-car remains significantly lower, suggesting low reliability in its predictions even at high confidence levels. The Precision-Recall (PR) curve in Figure \ref{fig:MaskPR_curve} evaluates the balance between precision and recall for each class across all thresholds. This figure is critical in understanding class-specific behavior, particularly in imbalanced datasets. The Private-car and Taxi classes maintain high precision and recall, whereas the Government-car class drops sharply. The mAP@0.5 value of 0.701 indicates good overall performance from the segmentation model in terms of detection quality.

\begin{figure*}[t]
\begin{center}
    \begin{subfigure}[b]{0.48\textwidth}
        \centering
        \includegraphics[width=\textwidth]{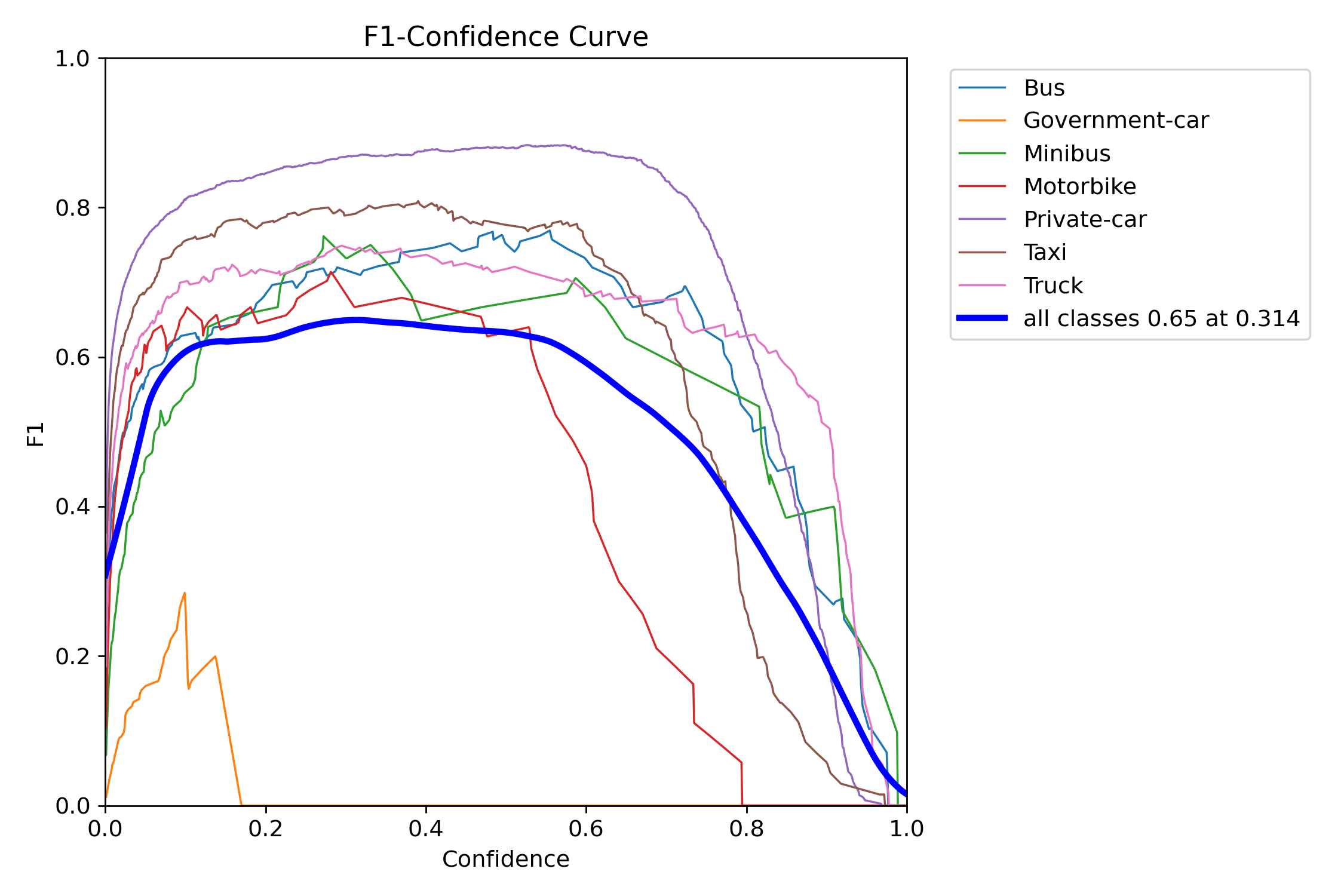}
        \caption{F1-Confidence Curve (Box)}\label{fig:BoxF1_curve}
    \end{subfigure} 
    \hfill 
    \begin{subfigure}[b]{0.48\textwidth}
        \centering
        \includegraphics[width=\textwidth]{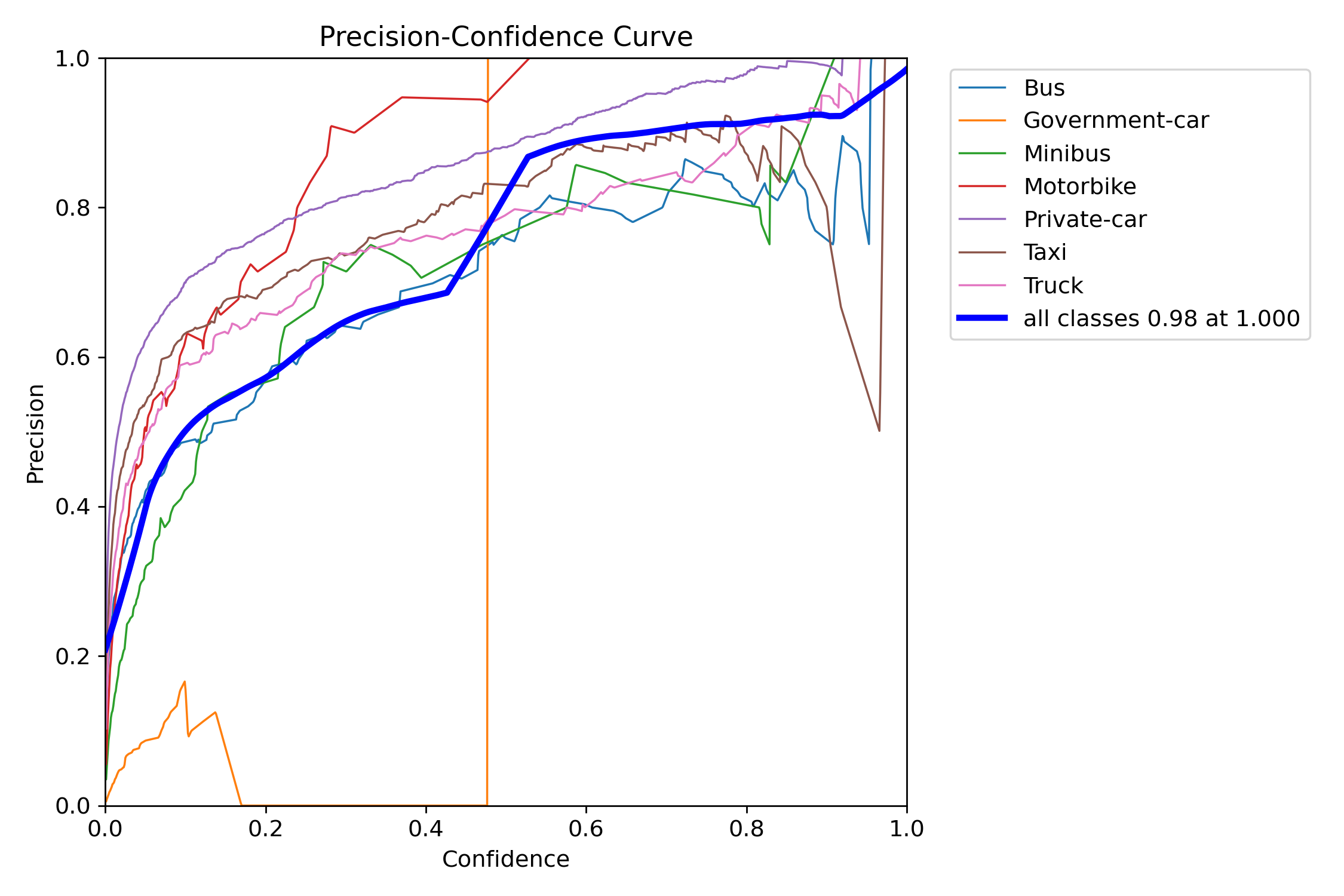}
        \caption{Precision-Confidence Curve (Box)}\label{fig:BoxP_curve}
    \end{subfigure}
    \vspace{0.5cm}
    \begin{subfigure}[b]{0.48\textwidth}
        \centering
        \includegraphics[width=\textwidth]{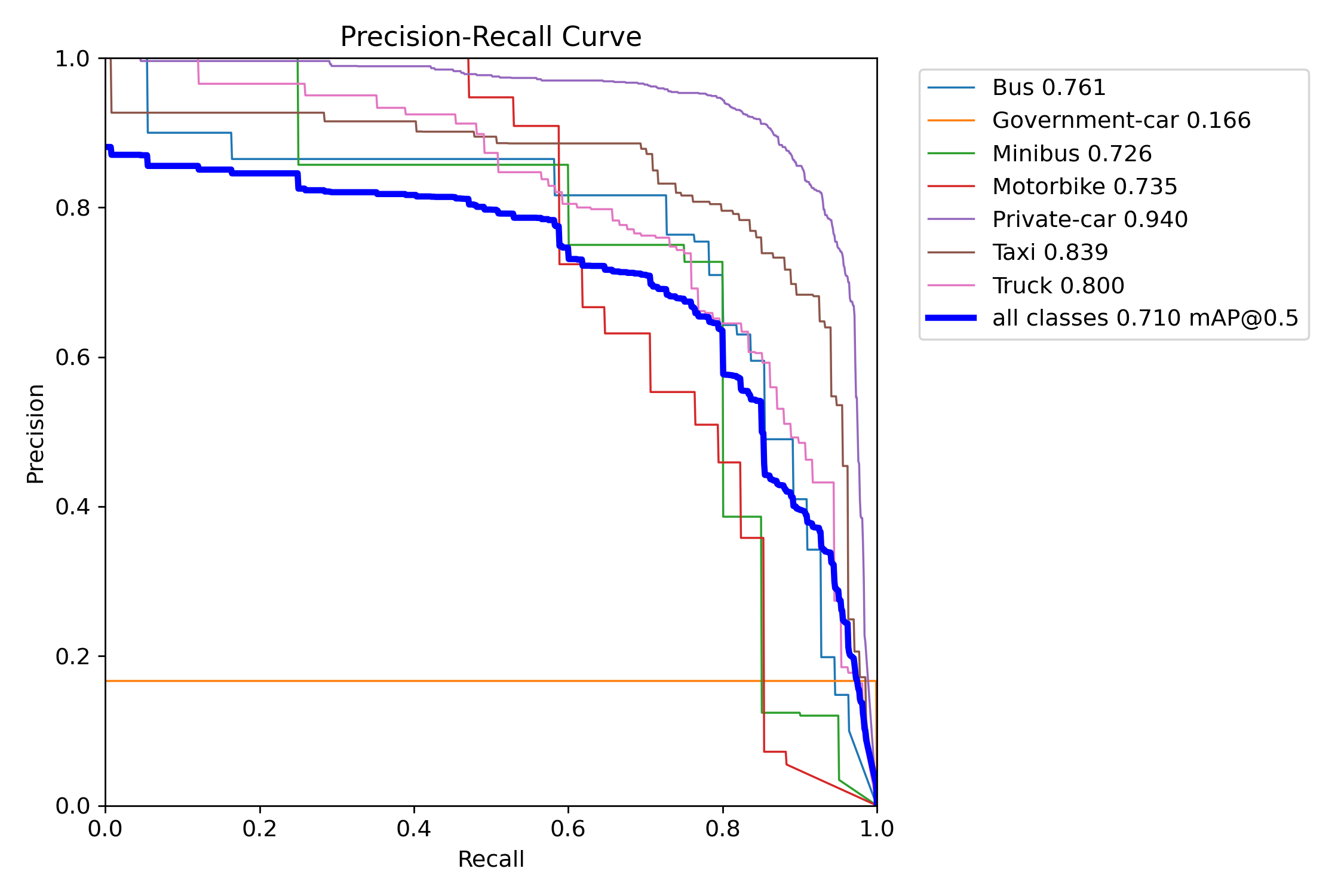}
        \caption{Precision-Recall Curve (Box)}\label{fig:BoxPR_curve}
    \end{subfigure}
    \hfill 
    \begin{subfigure}[b]{0.48\textwidth}
        \centering
        \includegraphics[width=\textwidth]{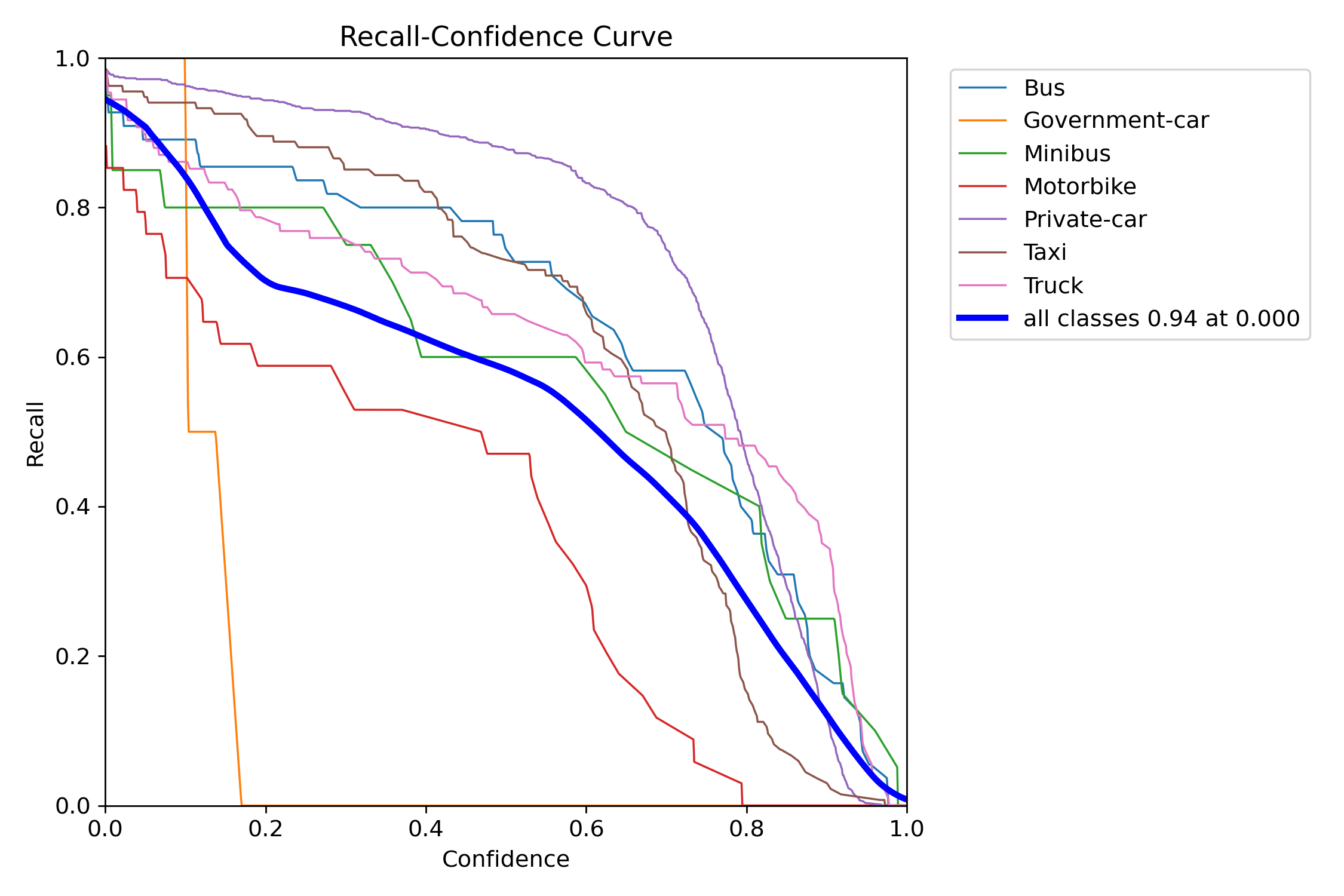}
        \caption{Recall-Confidence Curve (Box)}\label{fig:BoxR_curve}
    \end{subfigure}
    \caption{Comparison of various curves for the model (box). Figure~\ref{fig:BoxF1_curve} tracks the F1 score for bounding box detections over confidence thresholds. Best F1 for all classes is 0.65 at 0.314; Figure~\ref{fig:BoxP_curve} shows how precision varies across confidence levels. Precision peaks at 0.98 at a confidence of 1.000; Figure~\ref{fig:BoxPR_curve} shows precision vs. recall trade-offs based on bounding box predictions. The model reaches an overall mAP@0.5 of 0.710; and Figure~\ref{fig:BoxR_curve} shows how recall behaves as a function of confidence. The highest recall (0.94) is achieved at a confidence of 0.}
\end{center}
\end{figure*}

Moreover, Figure \ref{fig:MaskR_curve} highlights how recall drops as the confidence threshold increases. A recall of 0.93 is observed at the lowest threshold, meaning the model captures nearly all true positives. However, this also implies more false positives at low confidence. As the threshold increases, fewer detections meet the confidence requirement, and recall decreases. Private-car and Taxi maintain higher recall over wider ranges, while Government-car fails early. Furthermore, Figure \ref{fig:BoxF1_curve} shows curve functions similarly to the mask-based F1 curve but is derived from bounding box predictions instead. The optimal balance between precision and recall occurs at a confidence level of 0.314. The results are consistent with the mask model, suggesting that bounding box predictions can match segmentation performance under well-optimized thresholds. Performance varies by class, with consistent F1 gains for Private-car and Taxi classes. Additionally,  Figure \ref{fig:BoxP_curve} shows precision increasing with confidence, peaking at 0.98 at full confidence. High precision at 1.0 indicates the model is extremely accurate when it is most certain. However, the trade-off is fewer detections. The Motorbike and Private-car classes achieve strong precision across the range, while the Government-car class underperforms significantly.

While Figure \ref{fig:BoxPR_curve} reflects the model’s effectiveness in identifying true positives while minimizing false positives. A mean average precision of 0.710 at IoU threshold 0.5 shows that the bounding box detector performs slightly better than the segmentation model. Classes like Private-car and Taxi show near-ideal curves, while the sharp drop for Government-car reveals unreliability in that category. The plot in Figure \ref{fig:BoxR_curve} mirrors the mask-based recall curve, indicating that bounding box recall also starts high (0.94) and decreases as confidence increases. It shows the model’s tendency to recall more objects at lower thresholds but risk false positives. The curves help determine an optimal confidence threshold that balances detection volume and reliability.

\begin{figure*}[t]
\begin{center}
    \begin{subfigure}[b]{0.48\textwidth}
        \centering
        \includegraphics[width=\textwidth]{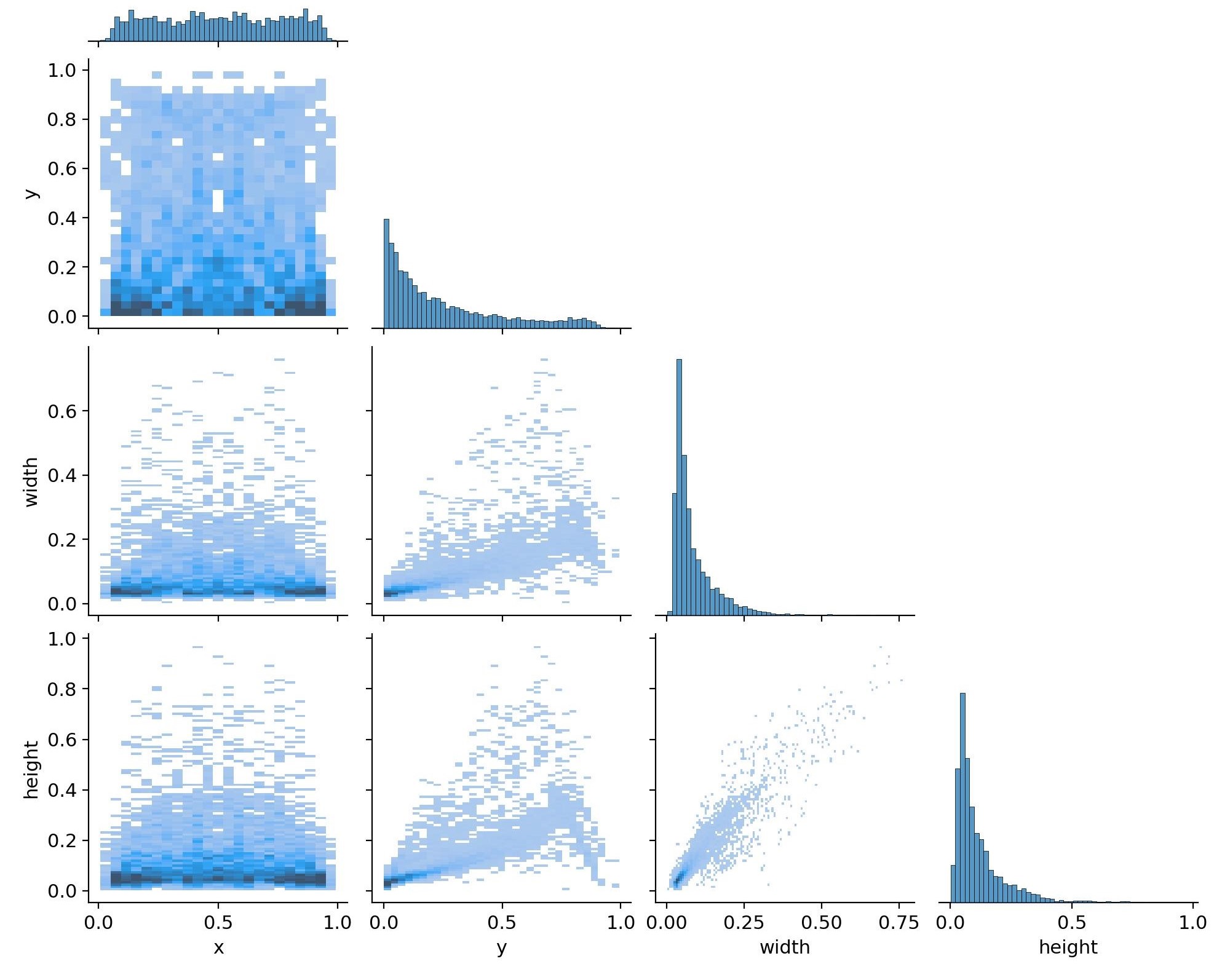}
        \caption{Pairwise Distribution of Normalized Bounding Box Parameters.}
        \label{fig:labels_correlogram}
    \end{subfigure} 
    \hfill 
    \begin{subfigure}[b]{0.48\textwidth}
        \centering
        \includegraphics[width=\textwidth]{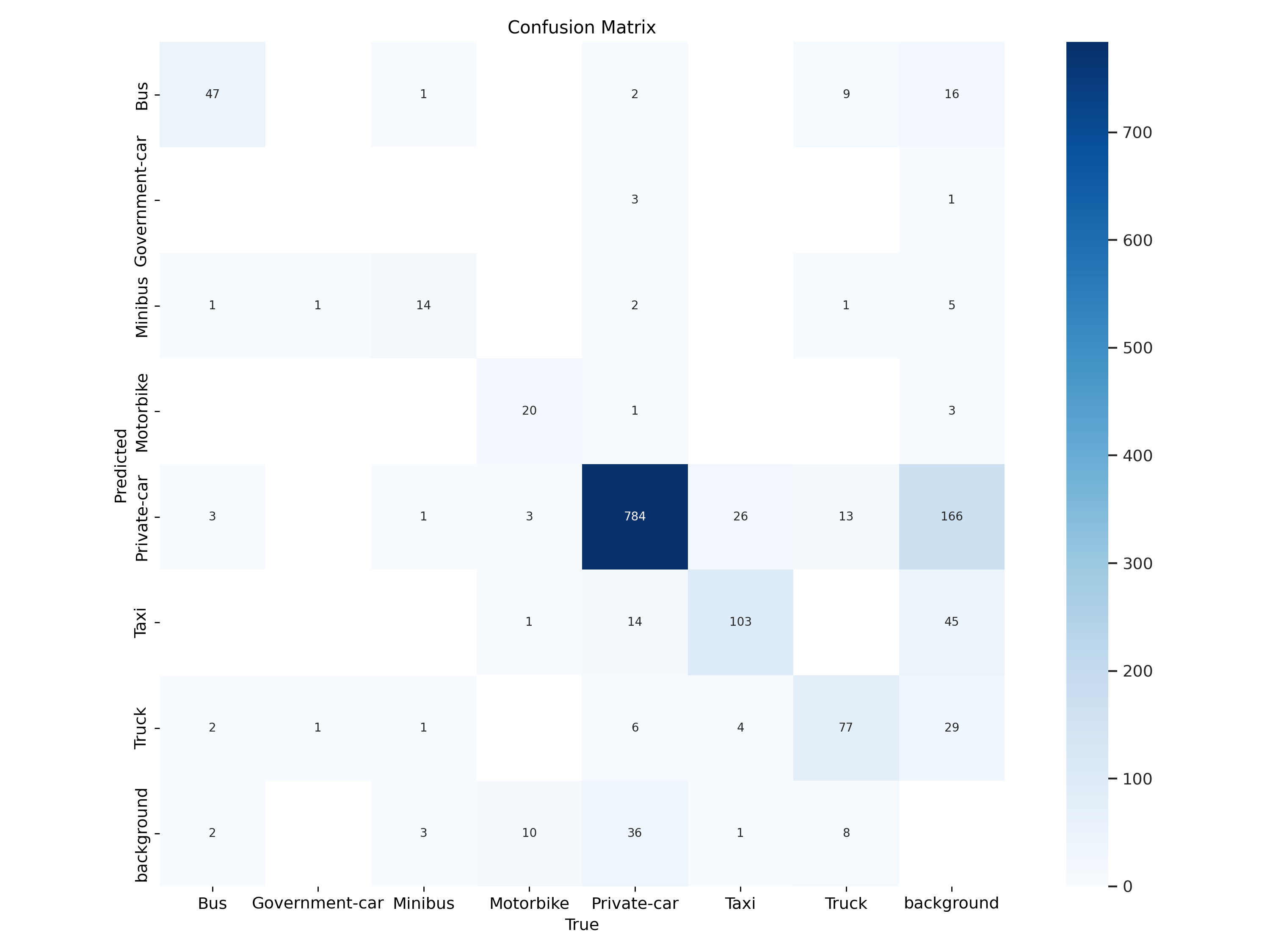}
        \caption{Confusion Matrix of Predicted vs. Actual Class Labels (Raw Counts).}
        \label{fig:confusion_matrix}
    \end{subfigure}
    \vspace{0.5cm}
    \begin{subfigure}[b]{0.48\textwidth}
        \centering
        \includegraphics[width=\textwidth]{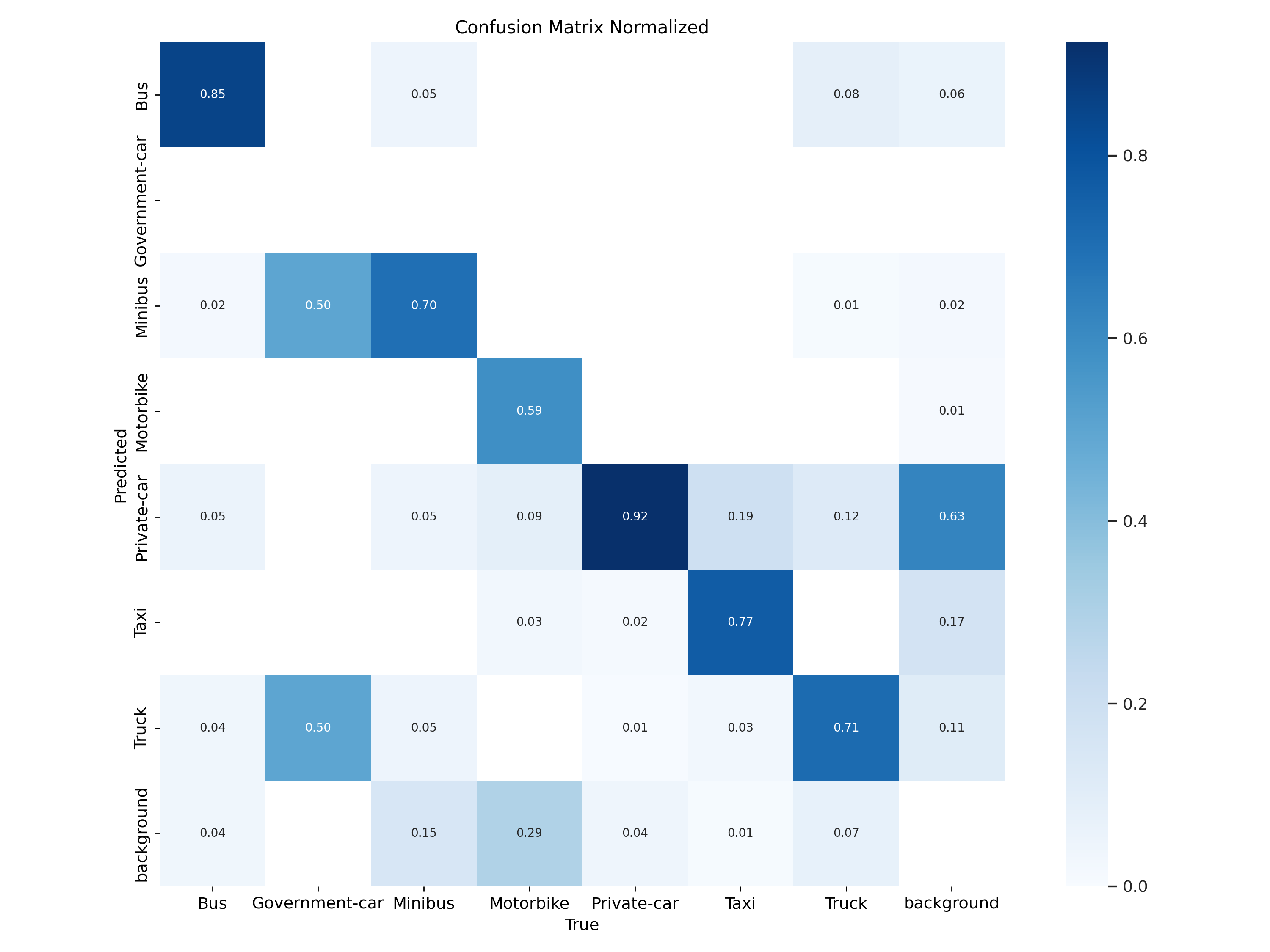}
        \caption{Normalized Confusion Matrix Showing Proportional Class Prediction Accuracy.}
        \label{fig:confusion_matrix_normalized}
    \end{subfigure}
    \hfill 
    \begin{subfigure}[b]{0.48\textwidth}
        \centering
        \includegraphics[width=\textwidth]{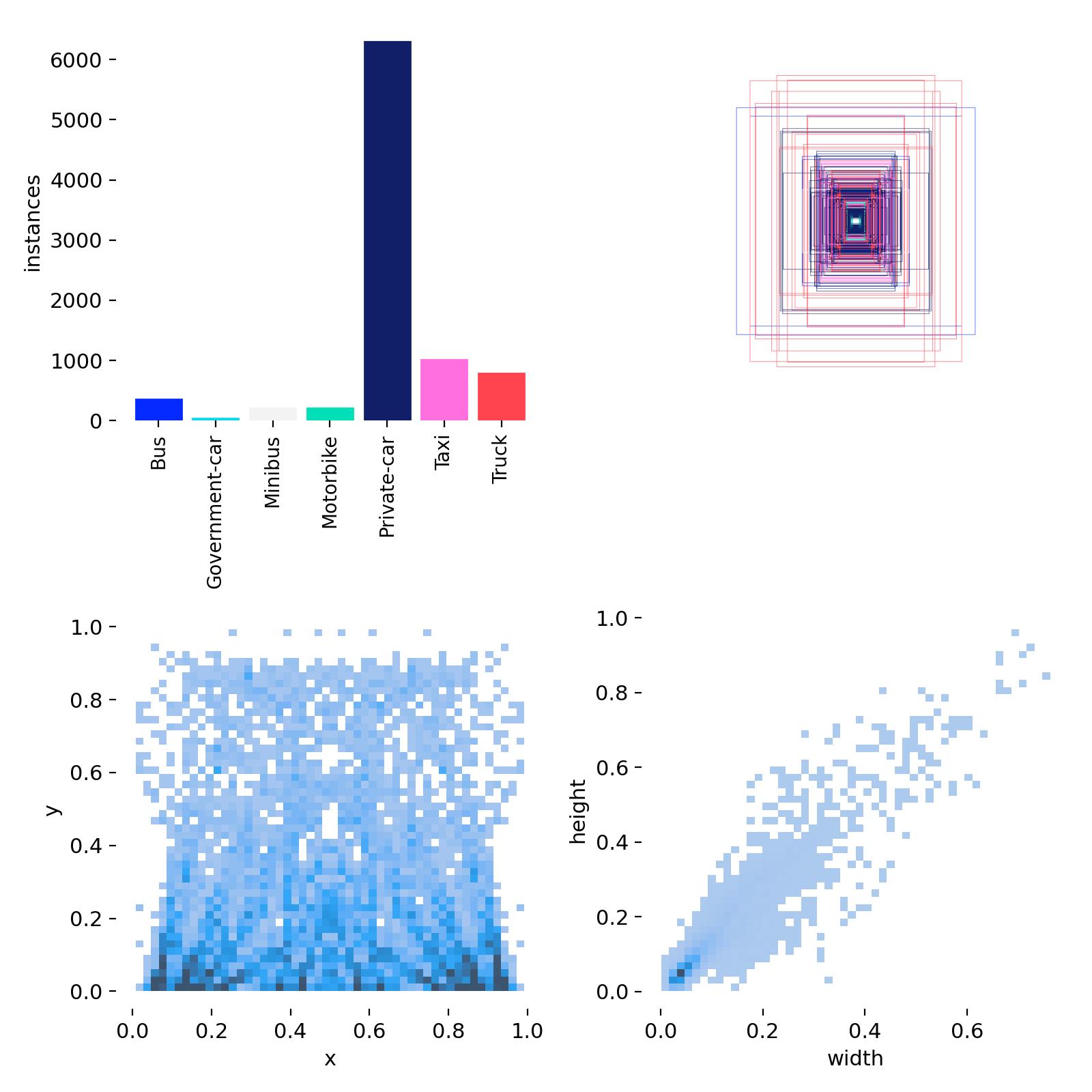}
        \caption{Label Frequency, Position Distributions, and Bounding Box Shapes.}
        \label{fig:labels}
    \end{subfigure}
    \caption{Visual representation of various evaluation outputs.}
\end{center}
\end{figure*}


The correlogram in Figure \ref{fig:labels_correlogram} illustrates pairwise distributions of bounding box attributes: \texttt{x}, \texttt{y}, \texttt{width}, and \texttt{height}, all normalized to fall within the range [0, 1]. The diagonal elements show individual histograms, indicating the distribution of each variable. Notably, the \texttt{y} coordinate and \texttt{height} values are highly concentrated near 0, implying most detected objects are positioned toward the lower part of the frame and are relatively small in size. The off-diagonal scatter plots reveal correlations, such as a mild positive relationship between width and height. This plot is useful for diagnosing dataset imbalance and typical object positioning in traffic scenes.


The confusion matrix in Figure \ref{fig:confusion_matrix} displays raw counts of true versus predicted labels for all vehicle classes. The darkest cell on the diagonal represents the 'Private-car' class with 784 correct predictions, indicating that the model performs best on this class. However, there is noticeable confusion with other classes like 'Truck' and 'Background', evidenced by the off-diagonal values in those columns and rows. For instance, 166 background instances were misclassified as 'Private-car'. The Government-car and Minibus classes have relatively few correct predictions, which may suggest underrepresentation or visual similarity with other classes in the dataset.


Moreover, the normalized confusion matrix in Figure \ref{fig:confusion_matrix_normalized} provides a proportion-based view of classification performance across classes, with each row summing to 1. This representation is particularly helpful for assessing how well the model performs relative to the total number of actual instances per class. Diagonal dominance—such as 0.85 for 'Bus' and 0.92 for 'Private-car'—confirms the model’s strong predictive ability for these classes. In contrast, classes like 'Minibus' and 'Motorbike' show more distributed predictions, highlighting difficulty in distinguishing them. The matrix also shows systematic confusion between visually or contextually similar categories such as 'Taxi' and 'Private-car'.

\begin{figure*}[h]

    \centering
    \includegraphics[width=\textwidth]{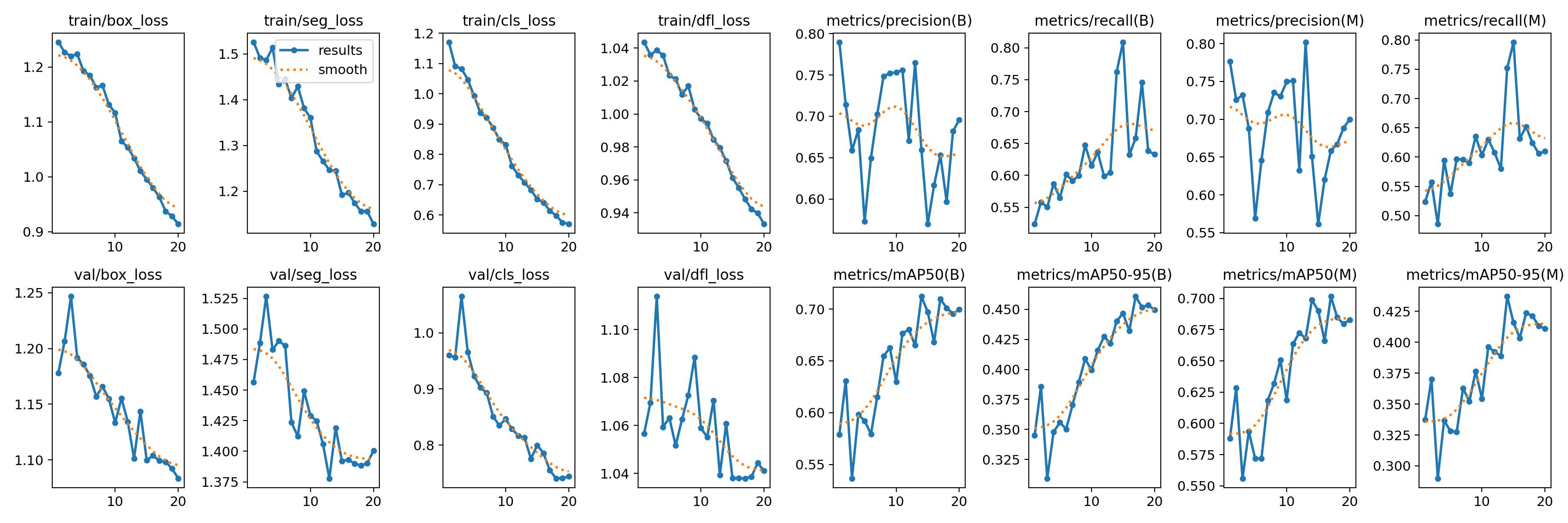}
    \caption{YOLOv8 Training and Validation Metrics Across 20 Epochs}
    \label{fig:results_metrics}

\end{figure*}

On the other hand, Figure \ref{fig:labels} combines several subplots to visualize data distribution and annotation geometry. The bar plot in the top-left shows the number of instances per class, revealing a significant class imbalance—'Private-car' dominates the dataset, followed by 'Taxi' and 'Truck'. In contrast, classes like 'Government-car' and 'Minibus' are underrepresented, which may impact model training and evaluation. The scatter plots below show the spatial distribution of object centroids (\texttt{x}, \texttt{y}) and the size relationships between \texttt{width} and \texttt{height}. Most objects are concentrated at the bottom-center of the image with small dimensions. The top-right plot stacks normalized bounding boxes to provide a visual sense of object scale and alignment across the dataset. Furthermore, Figure \ref{fig:results_metrics} displays a comprehensive overview of the YOLOv8 model’s training and validation progress across 20 epochs. The top row shows the training losses for bounding boxes (box\_loss), segmentation masks (seg\_loss), class predictions (cls\_loss), and distributional focal loss (dfl\_loss). All losses show a steady downward trend, indicating effective model optimization during training. The second row presents validation losses for the same components. While generally decreasing, these curves show more fluctuation due to variability in unseen validation data. Nonetheless, the declining pattern suggests that the model generalizes well. The third row displays precision and recall metrics for both bounding boxes (B) and masks (M). Recall generally improves over time, while precision shows more noise, highlighting sensitivity to false positives. Despite this, both metrics indicate consistent improvement. The final row includes mAP@0.5 and mAP@0.5:0.95 metrics for both box and mask predictions. These show a strong upward trend, with the box mAP@0.5 reaching approximately 0.70 by epoch 20, and mAP@0.5:0.95 nearing 0.45. These metrics confirm increasing detection and segmentation performance over the course of training.

\begin{figure*}[h]

    \centering
    \includegraphics[width=\textwidth]{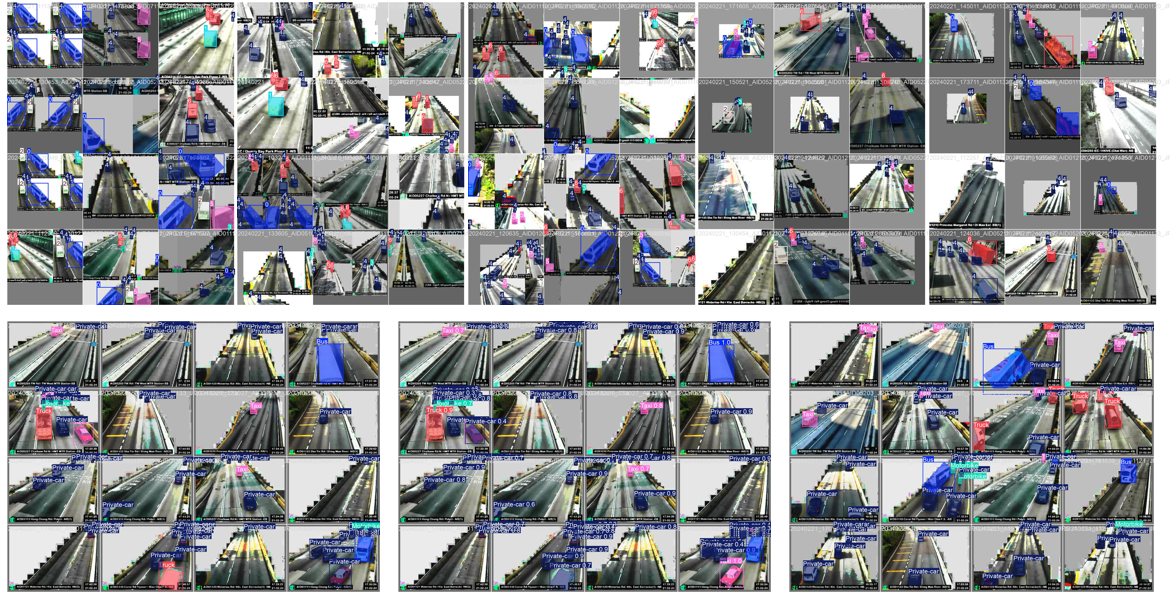}
    \caption{Visual output of YOLOv8 model during training and validation batches. The top half displays randomly sampled training batch predictions, while the bottom half shows model predictions on validation batches.}
    \label{fig:batch_outputs}

\end{figure*}

Figure~\ref{fig:batch_outputs} illustrates qualitative outputs from the YOLOv8 model during training and validation stages. The top row of the figure shows sample predictions from various training batches. These samples are randomly drawn during training to monitor the model’s performance in real time. Each detected object is annotated with its predicted class label and confidence score, visualized using colored bounding boxes corresponding to each vehicle type. The model demonstrates increasing accuracy over time in segmenting and classifying vehicles such as private cars, buses, taxis, and trucks. While the bottom row displays predictions from the validation dataset. These images are used exclusively to evaluate the model's ability to generalize to unseen data. The consistency of detections and correct labeling across varied lighting, angles, and occlusions suggests good generalization. Although most predictions are accurate, some minor errors and misclassifications—such as confusion between trucks and private cars—can still be observed. These qualitative results complement the quantitative performance metrics and visually confirm the model's effectiveness in multi-class vehicle detection tasks under real-world traffic conditions.

\begin{figure*}[h]

    \centering
    \includegraphics[width=\textwidth]{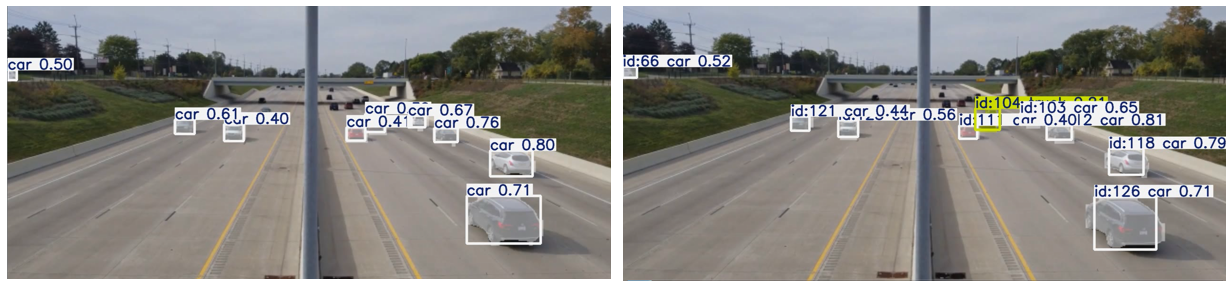}
    \caption{Proposed Pipeline process of detection and tracking outputs on a sample video frame. The left image shows vehicle detections using YOLOv8, while the right image demonstrates multi-object tracking (MOT) with unique IDs assigned for each vehicle.}
    \label{fig:tracking_ocr}

\end{figure*}

Figure~\ref{fig:tracking_ocr} presents a visual comparison between object detection and object tracking on a single video frame from a highway traffic sequence. The left image shows the output of the YOLOv8 detector applied to one frame, where vehicles are detected and annotated with bounding boxes and class confidence scores. However, no vehicle identities are maintained between frames at this stage. The right image illustrates the application of a Multi-Object Tracking (MOT) algorithm that assigns persistent identity labels (IDs) to each vehicle across frames. This enables the system to keep track of individual vehicles as they move through the scene. The tracked bounding boxes, marked with both class and ID, are then used to crop vehicle images accurately and consistently across frames. These cropped images serve as input to the OCR pipeline, which extracts license plate information from each uniquely tracked vehicle. This step is essential for ensuring that each license plate is only processed once and associated with the correct vehicle, supporting accurate carbon emission estimation and statistical vehicle analysis in real-time traffic monitoring scenarios.

\section{Conclusion} \label{sec:conclusion}

This study presents a novel and practical framework that integrates state-of-the-art deep learning techniques—specifically YOLOv8 for object detection and a specialized OCR model for license plate recognition—into a real-time system for estimating carbon emissions from road traffic. The fusion of these technologies enables the accurate identification and classification of individual vehicles in live traffic environments and provides a scalable approach to environmental monitoring and emissions analysis. Moreover, the use of YOLOv8 as the core detection engine in this pipeline demonstrates clear advantages over traditional methods. YOLOv8’s robust architecture offers high-speed and high-accuracy detection of multiple vehicle classes, even in complex urban environments. Its ability to output both bounding boxes and segmentation masks enhances the localization precision, which is critical for isolating vehicles for further processing. Building upon this, the system applies a multi-object tracking mechanism to ensure consistent tracking of vehicles across video frames, preventing duplication and maintaining the integrity of vehicle data. Each tracked and cropped vehicle instance is then passed through a deep OCR model capable of extracting license plate numbers under varied real-world conditions. The OCR model, designed with convolutional layers for spatial feature extraction and recurrent layers for character sequence decoding, achieves reliable plate recognition even in cases of partial occlusion, motion blur, or poor lighting. The recognized license plate numbers serve as keys to query a remote vehicle database API, which returns metadata including vehicle type, fuel category, engine size, and carbon emission coefficients. This step is crucial, as it transforms basic detection into actionable environmental insight. With this information, the system calculates the estimated carbon emissions for each vehicle using standardized emission factors, contributing to a granular and accurate emissions profile for the monitored area. What makes this approach particularly novel is its ability to move beyond conventional traffic monitoring to enable individualized, real-time emission estimation at scale. Rather than relying on averaged or aggregated data, the proposed system delivers per-vehicle carbon emission assessments by integrating cutting-edge visual recognition models with external data systems. This opens new possibilities for intelligent traffic systems, environmental regulators, and policymakers who seek to measure, control, and ultimately reduce emissions on a vehicle-by-vehicle basis. The framework aligns with global sustainability initiatives and supports net-zero targets by providing an innovative digital infrastructure for emissions accountability. It can be deployed in smart city environments to generate live environmental reports, enforce green transportation policies, and identify high-emission vehicle categories that may benefit from targeted interventions. In summary, the integration of YOLOv8 and OCR technologies within this real-time, vision-based carbon emission monitoring pipeline represents a significant step forward in applying artificial intelligence to environmental sustainability. It demonstrates how advanced computer vision can be harnessed not only to understand urban mobility but also to inform strategies toward a more sustainable, low-carbon future.

\authorcontributions{Formal analysis, investigation, conceptualisation, methodology, validation, software, supervision, project administration [NN, AA]; data curation, visualisation [AA]; writing---original draft [NN, AA, MA, AH]; resources, writing---review and editing [NN, AA, MA, AH, SR]. All authors have read and agreed to the published version of the manuscript.}

\funding{This research received no external funding.}

\institutionalreview{Ethical review and approval were waived for this study because the datasets used for experiments are from open access articles published under a CC-BY license, which allows for unrestricted use, distribution, and reproduction in any medium, provided the original work is properly cited, and all protocols for responsible data handling and usage have been followed.}

\dataavailability{The dataset(s) used in this study are publicly available from two Roboflow projects: Road-Traffic Computer Vision~\cite{mmsseg-dataset} and Number Plates Computer Vision~\cite{number-plates-dataset}. Both projects are published under CC BY 4.0 license, allowing reuse and adaptation with proper attribution, hence the citation.} 

\acknowledgments{This work inspired aspects of the methods developed for the British Academy and Nuffield Foundation project on Understanding Communities (Grant number FR-000023474). We thank Trinity College Dublin and  thank the School of Engineering and Sustainable Development at De Montfort University and  Computer Science Department at Edge Hill University and University of Technology, Iraq for providing resources and support.}


\bibliographystyle{unsrt}


\end{document}